# Rashid: A Cipher-Based Framework for Exploring In-Context Language Learning

**Niyati Bafna[1], Ryan Soh-Eun Shim[2], Barbara Plank[2],**
**David Yarowsky[1], Hale Sirin[1]**

[1]Johns Hopkins University, Center for Language and Speech Processing;

[2]MaiNLP, Center for Information and Language Processing, LMU Munich

nbafna1@jhu.edu

## Abstract

While there is growing interest in in-context language learning (ICLL) for unseen languages with large language models, such languages usually suffer from the lack of NLP tools, data resources, and researcher expertise. This means that progress is difficult to assess, the field does not allow for easy large-scale experimentation, and findings on ICLL are often limited to very few languages and tasks. In light of such limitations, we introduce a framework (Rashid) for studying ICLL wherein we reversibly cipher high-resource languages (HRLs) to construct truly unseen languages with access to a wide range of resources available for HRLs, unlocking previously impossible exploration of ICLL phenomena.[1] We use our framework to assess current ICLL methods for machine translation with SOTA evaluation tools and manual analysis, explore new directions leveraging otherwise expensive resources for improving ICLL, and test ICLL strategies for rich downstream tasks beyond machine translation. These lines of exploration showcase the possibilities enabled by our framework, as well as providing actionable insights regarding current performance and future directions in ICLL.[2]

## 1 Introduction

The task of processing unseen or extremely low-resource languages with large language models (LLMs), first introduced by Tanzer et al. (2023), holds the potential to make important strides in extending the benefits of LLMs to the long tail of languages. This is typically done by augmenting the prompt context with information about the language, using materials such as lexicons, grammar books and other tools for linguistic processing, as explored by several subsequent works (Zhang et al., 2024b, 2025; Hus and Anastasopoulos, 2024).

However, progress in the field of in-context language learning (ICLL), by its nature, is hindered by resource scarcity. The unseen or truly low-resource languages of relevance to ICLL are the very languages for which we lack a) quality evaluation strategies including access to human expertise, b) high-quality resources for cheaply exploring a range of new strategies without significant investment, and c) task data from frontier benchmarks, for testing the performance of these strategies in various application scenarios. Further, insights from these setups regarding language learning are often confounded by the model's differing existing capabilities in various low-resource languages.

In this work, we introduce Rashid, a general-purpose framework for systematically assessing and exploring the phenomenon of ICLL. We construct truly unseen languages by reversibly ciphering high-resource languages (HRLs), preserving underlying linguistic characteristics, and providing us access to a wide array of available resources in these languages for experimentation. See Figure 1 for an illustration of the framework. Crucially, this allows us to bypass the above problems of resource scarcity and factors confounding insights, enabling sophisticated evaluation as well as the assessment of techniques otherwise requiring potentially expensive resource development for true LRLs. We make the following contributions:

**Framework** We construct a sandbox for exploring in-context language learning for languages guaranteed to be unseen, with access to the range of NLP tools and resources available for HRLs.

**Assessing status quo** For the first time, we use SOTA evaluation techniques and manual analysis to assess current methods for ICLL for machine translation (MT) in both directions. We find that current

---

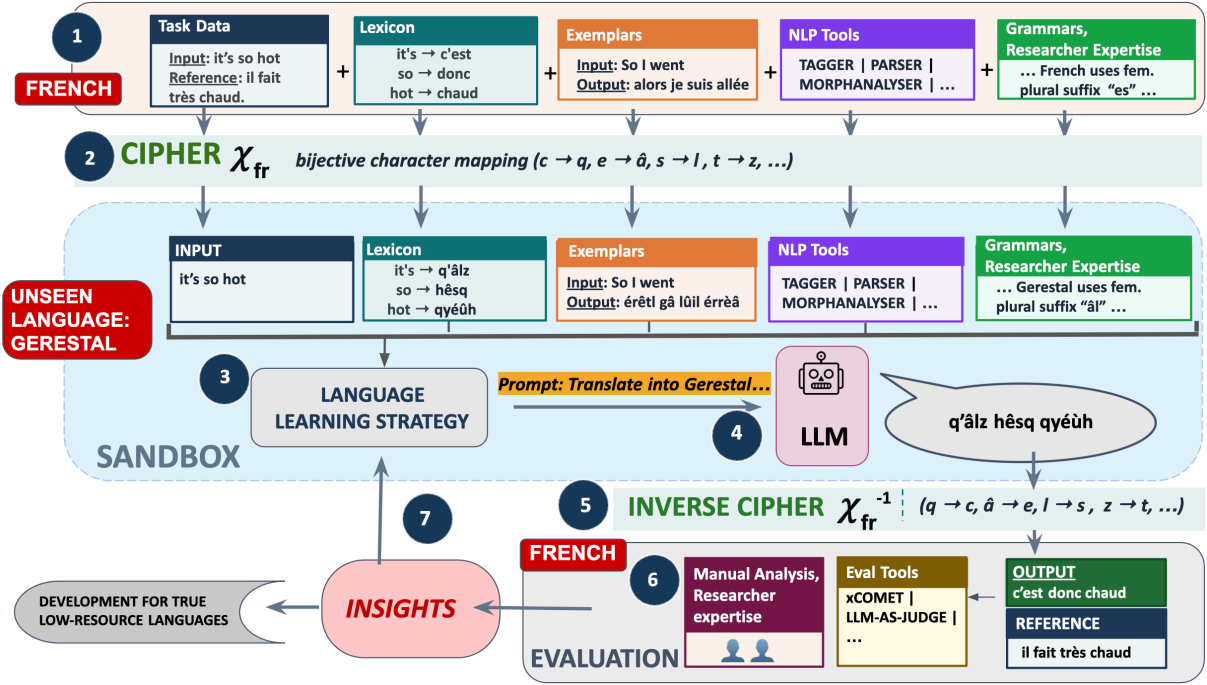

Figure 1: RASHID, our framework for exploring in-context language learning for arbitrary tasks. We show an example task (MT, eng→). (1) Starting with a range of NLP tools, data resources, and researcher expertise in a high-resource language (French), (2) we apply a bijective character-level cipher $\chi_{fr}$ over HRL text, allowing us to construct an "unseen" language ("Gerestal") with access to the same range of materials. (3) We test several language learning strategies drawing from relevant resources for (4) in-context language learning with an LLM. (5) Model outputs can be reverted to the HRL with $\chi_{fr}^{-1}$ when relevant. (6) This enables sophisticated evaluation available for HRLs as well as the (7) iteration over new strategies from resulting insights.

strategies show intended benefits for *comprehension* of unseen languages (→eng), although they remain ineffective for *generation* (eng→), and that outputs remain low-quality for both directions. We present a schema of challenges for this paradigm of language learning.

**Exploring new directions** We explore two research directions for improving ICLL. We find that near-perfect lexicon coverage would be helpful, and that cascading through a related high-resource language for generation makes considerable progress on the problems of syntactic coherence and translation of idiomatic language. This promotes the development of rich parallel materials between related languages rather than with English.

**Evaluation on downstream tasks** We test current ICLL approaches on three tasks beyond MT. We find that added materials may not be directly beneficial for downstream tasks, but that a cascaded approach (MT with ICLL → task-solving) improves performance across the board. This underlines the need for task-oriented exploration of ICLL.

We emphasize that we do not intend this work to replace valuable studies on true LRLs. Instead, we showcase the power of our framework in providing a supplementary sandbox for systematic and broad experimentation in cases where this is hindered by resource scarcity, and for guiding expensive resource creation.

## 2 Previous Work

**In-context Language Learning** Tanzer et al. (2023) first showed MT performance on an "unseen", language, Kalamang, by including word meaning, bitext examples, and a grammar book in the context of the prompt. This was followed by work that explores this phenomenon as aided by several such tools and materials. This includes using lexicons, morphological tools, such as morphological analysers and interlinear glossed text to augment the prompt with relevant information (Elsner and Needle, 2023; Zhang et al., 2024b; Lu et al., 2024; Zhang et al., 2025; Ramos et al., 2025), as well as syntactic information, such as curated syntax rules and grammar documentations (Zhang et al., 2024a; Hus and Anastasopoulos, 2024; Hus et al., 2025). Pei et al. (2025) and Aycock et al. (2024) compare the usefulness of different materials for Manchu and Kalamang respectively. Li et al. (2025) compares few-shot ICL with provided word meanings with fine-tuning for low-resource languages. These studies largely focus on machine translation into English evaluated with string-based metrics, often involve expensive resource curation sometimes limiting experiments to a single language, and do not provide manual analyses or usability assessments of outputs, presumably due to hard-to-find resources and expertise in studied languages.

**Ciphering in NLP**  has been used for varied reasons besides decipherment studies (Knight et al., 2006; Berg-Kirkpatrick and Klein, 2013; Kambhatla et al., 2018; Shen and Smith, 2025). K et al. (2020) used a ciphered version of natural language data to study structural linguistic transfer as opposed to surface-similarity-based transfer in BERT-like models (Devlin et al., 2019). In the context of LLM safety and jailbreaking, Yuan et al. (2023) explore LLM capability to comprehend and generate text in conventional ciphers (such as shift-based) given instructions to do so, and Guo et al. (2025) show that LLMs struggle to provide reasoning in well-known ciphers, given instructions. Substitution ciphers have also been used to remove LLM reliance on pretraining data while testing ICL abilities (Fang et al., 2025; Pei et al., 2025). As far as we know, we are the first to use substitution ciphers as a strategy for overcoming the challenges of low-resource multilingual settings to enable systematic experimentation in these settings.

## 3 Constructing unseen languages: Ciphers

Our goal is to construct truly unseen languages that have a wide range of tools and resources available for them. We obtain these languages by *ciphering* high-resource languages at the character-level in a reversible manner. We use a monoalphabetic substitution cipher: for each language $L$ with a character inventory $\Sigma_L$, we construct a permutation $\chi_L : \Sigma_L \to \Sigma_L$. Thus, $\chi_L$ is a bijection with an existing inverse $\chi_L^{-1}$.[3] We extend $\chi_L$ character-wise to strings; for any string $w = c_1 c_2 \ldots c_n \in \Sigma_L^*$,

$$\chi_L(w) = \chi_L(c_1)\chi_L(c_2)\ldots\chi_L(c_n).$$

We apply $\chi_L$ to text in $L$ in all materials $\{M_L^{lexicon}, M_L^{input}, \ldots\}$ in $L$ to obtain a ciphered language $L^\chi$ with corresponding materials $\{M_{L^\chi}^{lexicon}, M_{L^\chi}^{input}, \ldots\}$. We can decipher model outputs $y_{L^\chi}$ in $L^\chi$ back into $L$ when relevant by applying $y_L = \chi^{-1}(y_{L^\chi})$ (shown in Figure 1). Each ciphered language is referred to by an invented name and presented to the model as a "new" language. See more details about the ciphers and examples of ciphered text in Appendix A. Our cipher has the following key characteristics.

**Reversibility**  Ciphered text should be losslessly revertible to the original language, allowing decipherment of model outputs for evaluation.

**Obscuring underlying language**  The model should process $L^\chi$ as an independent unseen language, and should not be capable of deciphering it under task constraints. Our character substitution cipher yields $\approx |C|!$ maps for a character inventory $C$, and would require cryptanalytic expertise, a considerable amount of ciphertext data, and cryptanalytic intent, to decipher (Shannon, 1949; Knight et al., 2006). We validate that LLMs are unable to undo the cipher even with explicit step-by-step instructions via a `decipher-then-translate` test (§ A.2); thus, they are forced to rely on language learning to perform the task. We further provide a sanity check baseline in § 5), showing that models perform as expected for unseen languages.

**Maintaining underlying natural language properties**  Our character-level cipher maintains all structural linguistic properties of the base language and its language family apart from surface forms, including usage patterns of (ciphered) words and morphemes, and morphosyntactic and semantic properties.[4] Evaluating on ciphered languages thus lends insights on hypothetical truly unseen languages with similar properties. We acknowledge caveats with this claim regarding the disruption of certain ortho-phonological properties like vowel harmony in § 9.

---

[3]We permute vowels, consonants, and diacritics when relevant separately among themselves, so that words in the ciphered language retain reasonable vowel usage and maintain orthographic conventions in that script.

[4]Named entities should generally be recognizable to the model even in unseen languages. We deal with this by always providing a glossary of ciphered named entities to the LLM, using named entity recognition over the original language text. We manually confirm that the LLM is able to trivially copy these into outputs and exhibit expected behaviour as over true unseen languages with respect to named entities.

## 4 Experimental setup

**Languages**  We work with ten mid-to-high resource languages covering seven language (sub)families and three scripts in order to assess the effectiveness of ICLL strategies across typologically diverse scenarios: German, Turkish, French, Spanish, Telugu, Hindi, Marathi, Vietnamese, Czech, and Polish. These include languages with various linguistic profiles in terms of morphology (e.g. Vietnamese is isolating, Polish is fusional, and Turkish is agglutinative) and syntax (e.g. Hindi is SOV while Spanish is SVO, Turkish and Marathi are freer word order unlike French).

**Models**  Our primary analysis is conducted with `GPT-5.1` (closed-source), and we additionally provide results with `Llama-3.1-8B-Instruct` (Grattafiori et al., 2024) and `Qwen-2.5-7B-Instruct` (Team, 2024) (open-source). We use default sampling hyperparameters for `GPT-5.1` and greedy decoding for the other models.

**Tasks**  We primarily work with MT and evaluate on the `WMT24++` dataset, which contains paragraph-level text across four domains (Deutsch et al., 2025). We describe our experiments on three more task datasets, `MMLU-ProX` (Xuan et al., 2025), `XNLI` (Conneau et al., 2018), and `XStoryCloze` (Lin et al., 2022), in § 8.

**Evaluation metrics**  While we are often forced to use string-based metrics for MT evaluation in low-resource settings, these have been shown to be much worse than semantics-based metrics in terms of correlation with human-perceived quality (Freitag et al., 2021, 2022). We accordingly report xCOMET (Guerreiro et al., 2024) and `GEMBA-MQM` (Kocmi and Federmann, 2023) in addition to chrF (Popović, 2015). The former is a trained neural metric. The latter represents increasingly popular LLM-based metrics where an LLM is prompted to list errors in a translation with associated category and severity. This list is used to derive a weighted score ranging from 25 to 0, lower being better.[5]

## 5 Current approaches

Previous work mentioned in § 2 uses four basic types of resources to augment the prompt: **lexicons** with word-level alignment, in-context task exemplars, **morphological information** or glosses per word including part-of-speech, and grammar books with the **syntactic profile** of the language. We assess the following approaches for both MT directions (→eng and eng→), including two baselines (`only-input`, `L-str`), and five methods of constructing the prompt for the LLM that use incrementally added materials. The materials are appropriately formatted by direction, e.g. word meanings provided as mapping from input language to the output language.

- `topline`: We compute the topline performance for the (unciphered) language, with a basic task prompt and no extra information.

- `only-input`: Zero-resource LLM baseline, consisting of simply the task prompt with ciphered language input. This produces unrelated text showing trivial performance, e.g. from copying the input, punctuation, or named entities (< 5 BLEU for all languages), and serves as a sanity check random baseline.

- `L-str` (non-LLM): Word-for-word translation using a lexicon, constructed by string concatenation of top lexical equivalents for every word in the input. We use fuzzy lexicon search for the input word with normalized edit distance as per previous work (Zhang et al., 2024b).

- L: Word meanings from a **lexicon** provided in the LLM context for every word in the input. We use the top $k = 2$ source matches per word with fuzzy matching, listing up to two equivalents per source match in the prompt if the lexicon lists multiple.

- LE: (Additionally) $E = 3$ input-output **exemplar** pairs, selected by BM25 similarity to the input from a held-out exemplar pool, in line with previous work (Zhang et al., 2024a).

- LELem: (Additionally) Lexicon matches for the lemmas of input words, obtained with Stanza lemmatizers (Qi et al., 2020). This mimics a scenario where we can build a **basic morphological lemmatizer** but not more complex tools for a new language.

- LELemM: (Additionally) **Richer morphological glosses** including part-of-speech tags and word features from Stanza. This typically includes plurality, person, case, tense-aspect-mood, and other such information.

- LELemMS: (Additionally) High-level **syntactic profile**, including 8 primary word order features and comments about its morphosyntactic typology based on WALS features (Dryer and Haspelmath, 2013). While previous works have used grammar books for this purpose, Aycock et al. (2024) shows that gains from this are largely due to bitext examples or lexical leakage in the grammar books. We instead use a manually constructed grammatical description as a controlled manner of providing syntactic information to the model.

We use ChiKhaPo lexicons from Chang and Bafna (2025), consisting of GATITOS (Jones et al., 2023) and PanLex (Kamholz et al., 2014).[6] See more details about all strategies and prompts used in the paper in Appendix C.

## 6 Assessment of current strategies

In this section, we discuss the MT performance of the current approaches described above, shown for both directions and three metrics in Figure 2.[7] In general, we see that →eng performance improves over baselines with incrementally added materials, and LELemMS performs the best of current strategies. This is consistent with previous literature. However, for both directions and all metrics, we see that LELemMS performance for each language is considerably below topline performance, which indicates a wide scope for improvement.

**Comprehension vs. generation** We see that while current approaches are helpful for →eng, they do not help much with eng→ performance, which does not show much improvement from the non-LLM word-for-word baseline L-str to LELemMS as per semantics-based metrics. This is not surprising for the following reasons. Firstly, generation is naturally harder than comprehension (Arivazhagan et al., 2019; Bugliarello et al., 2020); while we may achieve a decent extent of comprehension simply by understanding the meanings of words present in a sentence and re-ordering them in a logical way, generation unto even a passable extent requires grasp of the target language's lexical patterns, morphology, and syntax. Secondly, following from the above: current approaches do not provide sufficient detail for the model to be able to generate into a language. *This indicates the large scope for generation-specific strategy improvements.*

**Insights from sophisticated metrics** We can understand chrF as providing us the string-level view, xCOMET as the semantics-based view, and GEMBA-MQM as a perspective on usability with respect to error count of the outputs.[8]

For →eng, we see roughly corresponding overall increases in chrF and xCOMET scores for current approaches over baselines.[9] GEMBA-MQM scores show a decrease in at most 3 penalty points for a language for any approach: this can be interpreted as the model making making about half a "minor" mistake less in LELemMS as compared to only-input (as per § 4). Further, only 11.9% of LELemMS outputs show a score better than 25, which is the worst possible score, implying concrete improvements only for few test examples, rather than consistently. Using the error category listing per sample by GEMBA-MQM, we compute a category-specific score per approach, and find that most improvements come from fewer "accuracy" errors, although there are slight "fluency" improvements as well (see Appendix G for more details). This makes sense since the model is capable of fluent English, and only improves in terms of semantic faithfulness to the source.

For eng→, chrF shows some progress for the baseline L-str approach, whereas the other two metrics do not, presumably because chrF relies on string matching and is fooled by nonsensical word-for-word translations. GEMBA-MQM shows only 1% of LELemMS outputs above the worst score. *The above discussion underlines the value in better metrics beyond string-matching for a more reliable and fine-grained understanding of performance.*

**Qualitative benefits of individual strategies** In order to obtain a qualitative sense of the nature of improvements made by each incrementally added material, we conduct a comparative manual review of outputs for four languages (Hindi, Marathi, French, and Turkish), both directions, and five current approaches (40 settings) and 30 samples per setting.[10] We present a summary of our observa-tions here, with more detailed comments in Appendix H, and example outputs from each approach in Table 6. We find that exemplars serve to improve performance on particular input words and topically relevant exemplars reduce topic drift. We observe that lemmatizer availability improves lexicon recall over input words, that morphological information improves the grammatical category alignment in some outputs for comprehension, and that high-level syntactic information is able to guide LLM output ordering in simple phrases for generation. In general, we find that the benefits of added morphosyntactic information degrade for longer inputs, and are inconsistent across and within outputs. Importantly, we find that overall quality remains low, rendering the outputs unusable for practical purposes, in both directions.

---

[6] https://huggingface.co/datasets/ec5ug/chikhapo

[7] We provide results for Llama-3.1-8B-Instruct and Qwen-2.5-7B-Instruct in Appendix D, and show performance breakdown over domains in Appendix E.

[8] See Appendix F for a small-scale validity analysis of GEMBA-MQM scores.

[9] We find weak correlation between per-sample chrF and xCOMET scores (0.25 and -0.28 for →eng and eng→ respectively), confirming literature in this field that these measure different kinds of improvements (Rei et al., 2020).

[10] Authors are native or fluent in languages studied in all manual analysis.

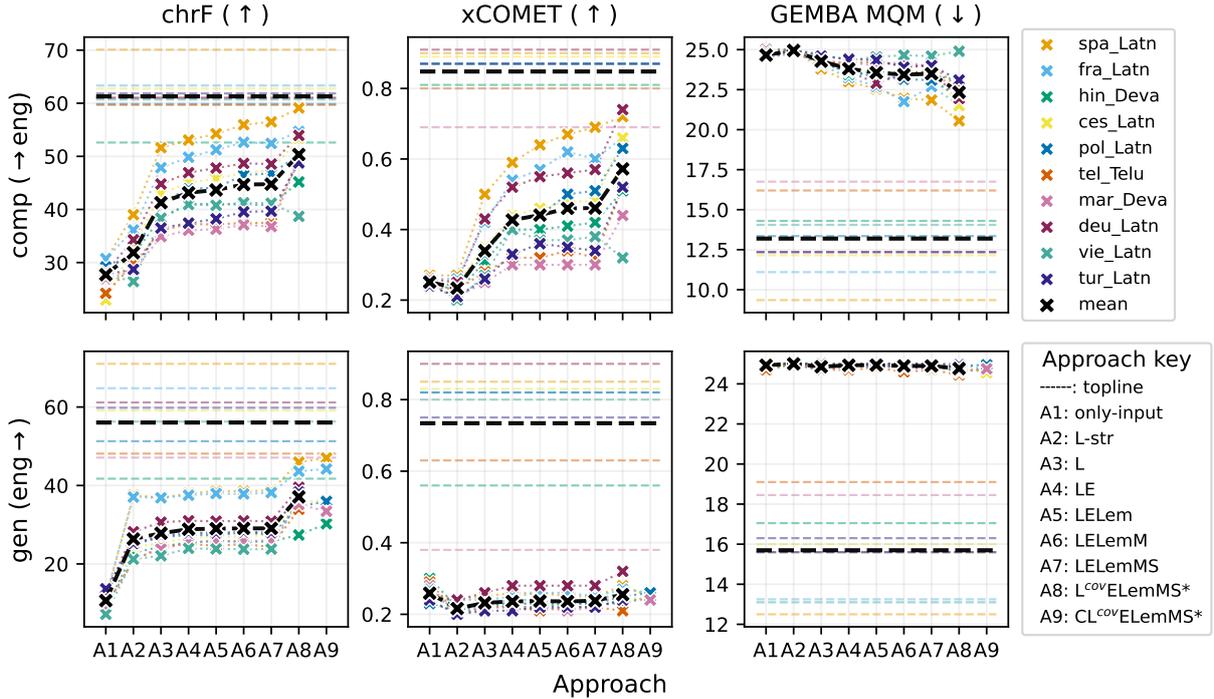

Figure 2: MT performance with `GPT-5.1` for →`eng` (top) and `eng`→ (bottom), evaluated with three metrics, for `topline`, baselines (`only-input`, `L-str`), current approaches (L- `LELemMS`), and new strategies (L$^{cov}$`ELemMS`, `CL`$^{cov}$`ELemMS`). Language means shown for all approaches except `CL`$^{cov}$`ELemMS`, which was evaluated for a subset of languages. *New strategies.

*Our quantitative and qualitative analyses show that a) incrementally added materials in current approaches yield intended improvements for language comprehension (→`eng`), b) approaches for language learning intended at language generation (`eng`→) are in their infancy c) translation outputs in both directions remain unusable and far behind* `topline` *in quality.*

**Characterizing challenges for ICLL** We further reviewed 100 examples in Turkish, French, and Marathi for `LELemMS` in both directions, and develop a schema of challenges that impede ICLL approaches within the current approach paradigm. See this schema in Table 1. These challenges broadly fall into two overlapping categories: a) coverage or accuracy failures in provided materials, and b) failures of the approach itself. *Our schema highlights the steep nature of this task, as well as inspiring directions for future strategies.*

## 7 Exploring new directions

Exploring solutions to the issues in Table 1 for truly low-resource languages may require considerable effort and investment in new resources, with no guarantee of good results. We now demonstrate how our framework can be used for fast preliminary exploration of research questions for boosting ICLL performance by investigating two such solutions. We show the quantitative results in Figure 2 (marked as new strategies). We also conduct a manual analysis of outputs for qualitative improvements similarly as described in § 6, and provide our findings from it in Appendix H, with example outputs in Table 6.

| Issue | Dir | Input | Output | Reference |
|---|---|---|---|---|
| †Word missing in lexicon/exemplars | mar→eng | त्वरित पिटाळले<br>*quickly dispersed* | escalated quickly | quickly dispersed |
| †*Fuzzy match returns irrelevant words | deu→eng | parasoziale Beziehungen<br>*parasocial relationships* | paranoid relationships | parasocial relationships |
| †*Wrong word sense provided by the lexicon | eng→fra | firm famous for sausage | ferme célébrité pour<br>*farm celebrity for*<br>saucisse<br>*sausage* | entreprise cèlèbre pour ses<br>*company famous for its*<br>saucisses<br>*sausages* |
| †*Morphological variant missing in lexicon | mar→eng | साधारण तासाभरापूर्वी<br>*about hour.one.ago* | the usual hour is | about one hour ago |
| *Word is provided in context but ignored | tur→eng | güneş panellerine<br>*solar panels.POSS.DAT*<br>göre<br>*according.to* | solar panels | According to the solar panels |
| *Morphosyntactic difference between languages causes failure | eng→tur | I hear nobody ask | ben hiç kimse sormak<br>*I no one ask*<br>duymak<br>*hear* | kimsenin sorduğunu<br>*anyone.GEN ask.PTCP*<br>duymadım<br>*hear.NEG.PST.1SG* |
| *Idiomatic language can't be reconstructed word-for-word | eng→fra | In other words | Dedans autre mots<br>*inside other words* | autrement dit<br>*in.other.words* |
| | eng→deu | we're back at it | wir andauern rucken a das<br>*we continue move at that* | also versuchen wir es<br>*so try we it*<br>noch mal<br>*again* |

Table 1: Schema of failures modes of current ICLL approaches affecting both comprehension (→eng) and generation (eng→), with examples. †: problem with materials, *: challenge of the approach paradigm.

## 7.1 Impact of better lexicon coverage

*RQ: What is impact of near-perfect lexicon coverage, including coverage over all morphological variants of nearly all input words?*

We use Google Translate over individual words as a proxy for a lexicon with near-perfect coverage of input words in running text ($L^{cov}$ELemMS). This means that each input word is covered by this hypothetical lexicon, including morphological variants and inflections, and that the obtained translation is often aligned morphologically to some extent (e.g. plurals would map to plurals).[11] However, note that problems like word-sense and phrase-level issues shown in Table 1 still remain.

We test this approach for all languages and both directions. We see that performance generally improves, especially for →eng, e.g. jumping by more than 10 xCOMET points for deu. This highlights the importance of lexicon coverage as the main supporting material aiding the model. For generation, the lack of morphosyntactic capabilities of the model still results in low-quality or nonsensical outputs.

Note that in practice, a lexicon such as this, covering all word forms in running text may involve building assistant technologies such as morphological parsers and morpheme-to-meaning dictionaries, depending on the morphological profile of the language. Further, performance even for →eng still lags considerably behind topline, given that issues such as idiomatic language translation and word sense still remain. *Our experiments indicate that a hypothetical lexicon with near-perfect coverage, outputs of the current paradigm would be improved for →eng, with the important caveat that other language learning challenges still need to be addressed for both directions.*

## 7.2 Using a related language pivot

*RQ: Given that related languages often have similar word order, can we mitigate mangled word order in eng→ by cascading through a related high-resource language to the target language?*

We first obtain a high-quality translation of the English input into the related high-resource language (RHRL) using Google Translate or GPT-5.1, and then use the RHRL as the input language for the LLM ($\text{CL}^{cov}$ELemMS). This means that we provide input word meanings and exemplars between RHRL and the ciphered LRL.[12]

---

[11]Given the mid-to-high-resource nature of our languages, Google Translate is of excellent quality and provides near-perfect coverage of input words.

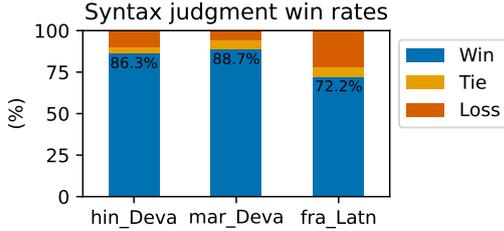

Figure 3: Syntax judgment win rates over 50 output pairs for $\text{CL}^{cov}$ELemMS vs $\text{L}^{cov}$ELemMS for 3 languages.

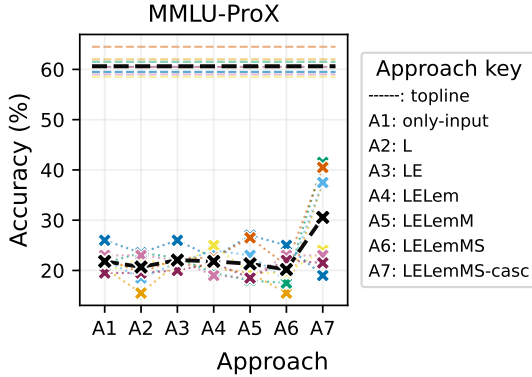

Figure 4: Results for `MMLU-Prox`

We test $\text{CL}^{cov}$ELemMS for eng→, for Marathi, Hindi, French, and Spanish, using Hindi, Marathi, Spanish, and French respectively as RHRLs, and compare against $\text{L}^{cov}$ELemMS. Quantitative metrics show slight gains, reflecting the fact that overall quality of outputs is comparable, given that many issues listed in Table 1 still remain. However, a manual win-rate evaluation targeted at naturalistic word order, shown in Figure 3, shows unanimous wins for $\text{CL}^{cov}$ELemMS. Qualitatively, we observe that word order is markedly improved especially for Hindi and Marathi, which are much more different from English in terms of word order than French. Further, lexical choice is often improved when the original English showed idiomatic phrasing, presumably because related languages often share semantic and lexical constructions, and word-for-word translation between them is more likely to succeed than from English. *These experiments pro-*

*mote building or leveraging parallel resources for LRLs with closely related high-resource languages rather than with English.*

## 8 Beyond MT: Other tasks

We evaluate current strategies on the `MMLU-ProX` (Xuan et al., 2025), `XNLI` (Conneau et al., 2018), `XStoryCloze` (Lin et al., 2022) tasks, containing multiple-choice questions for reasoning, entailment, and natural story continuation respectively. See accuracy results for `MMLU-ProX` in Figure 4. See Appendix I for similar results on the other two tasks and all prompts. We find that providing exemplars (LE) is useful, suggesting that task label alignment in the unseen language may play an important role. We see that providing further information such as word features and syntax overview (as in LELemMS) does not benefit performance, indicating that model may fail to leverage this information in a task setup rather than for translation. In general, performance lags far behind `topline` performance.

**Cascade** We also test a cascaded approach, that is often more performant for low-resource languages (Liu et al., 2025). That is, we first translate the ciphered language input into English using LELemMS, and then performing task inference in English (LELemMS-casc). Note that the information available to the model remains the same as in LELemMS. We find that LELemMS-casc performs the best for our tasks. This leading to several task-dependent questions regarding the role and trade-offs of using translation cascades as part of LLM pipelines for extremely low-resource languages.

*These initial findings motivate a task-oriented exploration of ICLL beyond the current focus on machine translation.*

## 9 Conclusion

Our work addresses the problem of resource scarcity that hinders the systematic and low-cost exploration of in-context language learning for unseen languages. We present RASHID, a framework which applies reversible character-level ciphers to high-resource languages to construct artificial truly unseen languages, unlocking access to the range of NLP tools and resources available for underlying high-resource languages. We demonstrate how this framework can be used for state-of-the-art evaluation of current approaches, exploration of new otherwise expensive strategies, and task-oriented investigation and insights. We aim to provide a grounded assessment of current techniques in in-context language learning, as well as enable further progress despite the many challenges of the field.

---

[12]In practice, lexicon coverage for this language pair may be much better or worse than between the target language and English. We avoid this confound by using near-perfect lexicon coverage, i.e. using Google Translate, and compare the results to a corresponding fair baseline.

## Limitations

**Unexplored directions** Our framework can be used to explore several research questions in language learning that we do not investigate in this work. It remains to explore the effectiveness of parametric methods (continued pre-training, task fine-tuning) at different data scales for adaptation to unseen languages, as well alternative inference techniques (chain-of-thought, various prompting styles). It remains to explore the effect of factors such as linguistic typology, relatedness to other high-resource languages, and script on language learning approaches, and to draw targeted conclusions regarding these, which we do not discuss in our work. Finally, there is naturally a vast range of different approaches to explore with regard to what type of materials can be leveraged for language learning to address the various challenges that we discuss in our work, and how they are presented to be used by LLMs. The goal of this work is to introduce our framework and demonstrate its flexibility in enabling cheap experimentation in all these directions. We leave it to future work interested in specific aspects to explore them in detail.

**Ciphering** While we claim that our ciphering technique maintains the linguistic properties of the underlying language, this has important caveats. Firstly, real unseen languages may have words in common or dialectal relationships with related languages, which our cipher does not currently emulate. This is intentional, allowing us to isolate "from-scratch" language learning from the effects of cross-lingual transfer, or of traces of language data in pretraining data. Our framework allows for controlled studies of these confounding effects, e.g. via partial ciphering for controlled token overlap, or other design choices modeling relatedness. We leave this to future work. Secondly, there may be particular ortho-phonological characteristics we may lose by random shuffling of the character inventory, such as vowel harmony, and the effect of tone. We leave it to future work to explore ciphers that can maintain these properties of interest for targeted studies of these phenomena. Finally, our current ciphering method would require adaptation to work for logographic scripts, such as Mandarin, where single graphemes are associated with meaning.

We note that our framework only requires that a cipher be perfectly reversible, or bijective. There are various ways to construct such ciphers other than monoalphabetic substitution ciphers as employed by us. For example, a cipher may have different and arbitrary input and output character inventories (i.e. mapping the Latin script to Devanagari instead of itself). Future studies modeling various scenarios may therefore design and employ arbitrary suitable ciphers as per research questions of interest as long as they satisfy this constraint.

**Modeling language-specific insights for true unseen languages** Our framework assumes that we can gain insights on true unseen languages by using artificial unseen languages, constructed using high-resource languages. This only holds if the target unseen language has similar properties to the considered HRLs, e.g. if they belong to the same language family, or share typological characteristics in terms of morphosyntax. However, if the target unseen has unusual properties that are not represented in the world's HRLs, our framework will not be capable of emulating the effect of those properties on language learning. More generally, our framework is not capable of providing language-specific insights for particular unseen languages and their niche typology and circumstances.

## Ethics Statement

Our framework uses ciphered high-resource languages as proxies for extremely low-resource or unseen languages, leveraging resources in the former that are unavailable for the latter. This risks the erasure of language-specific phenomena in such studies for languages that are already marginalized and under-represented if insights from this framework are not validated for real targeted languages. We emphasize in § 1 as well in this statement that we intend our framework to supplement rather than replace targeted studies for truly low-resource languages, in cases where experimentation on the latter is difficult or expensive due to resource scarcity. We do not work with any truly low-resource languages or language communities in this work, and acknowledge that the various language communities may have different goals than those of mainstream NLP (Bird, 2020).

## Acknowledgments

This work is partially supported by the AI and Advanced Computing Institute at Schmidt Sciences.

We would like to thank Aayush Mishra, Anjali Kantharuban, and Nathaniel Robinson for initial discussions, and Patricia Schmidtova, Kaiser Sun, Henry Li Xinyuan, and Avijit Thawani for feedback on the draft.

# A Ciphers

We describe our ciphering method here, and talk about its undecipherability given the context of experimental setup.

## A.1 Building cipher maps

Our studied languages include 7 languages using the Latin script, 2 languages using Devanagari script, and Telugu, which uses the Telugu script. In practice, it is common for extremely low-resource languages to develop a Latin-alphabet-based orthography if none exists (Jones and Mooney, 2017). We also include the Devanagari script, which is used by more than a hundred high- and low-resource languages and dialects (Eberhard et al., 2025), and Telugu script, which can be considered a low-resource script.

**Script-specific design** Seven out of ten of our languages use the Latin alphabet, although often with extra characters depending on the language. For each language, we construct disjoint consonant and vowel inventories. Each inventory is shuffled separately, so that a character maintains its class after the map. This is to avoid (for example) unrealistically long consonant sequences in the ciphered language.

The remaining three languages use the Devanagari and Telugu scripts. These scripts uses alternate graphemes (diacritics) for a vowel, with deterministic occurrence depending on the previous character (if any) in a word. We therefore use three classes: consonants, vowels, and vowel diacritics, and shuffle them separately. This allows us to maintain the orthographic conventions of the scripts.

**General design choices** We sample the cipher map once, and apply the same map over all input materials (i.e. input, lexicon entries, exemplars, etc.). This maintains any underlying correspondence between them.

Note that in our experiments, we choose to maintain the script of the source language as a design choice, in order to i) maintain a connection of the ciphered language with its language family ii) process various scripts. This is not necessary: it is possible to construct a bijective random map between equal-sized subsets of any two character inventories. This can be used to study script effects in a controlled manner in the future.

**Examples of ciphered text** We provide some examples of ciphered text in Table 2.

## A.2 Discussing indecipherability of ciphers by LLMs

Several past work has discussed that substitution ciphers such as ours have an exponential keyspace and are only decipherable using statistical frequency analysis given considerable amounts of ciphertext data (Shannon, 1949; Knight et al., 2006; Nuhn and Ney, 2013). Work specifically in the context of LLM ability to work with ciphers shows that even when cipher rules are provided to LLMs for much simpler types of shift-based ciphers, they are not consistently able to engage in reasoning in ciphertext (Guo et al., 2025).

We provide an explicit `decipher-then-translate` test, and find that ciphers built in the above manner are secure in the context of our prompting setup.

### A.2.1 Decipher-then-translate test

**Experiment setup** We maintain our prompting setup with the relevant ciphertext material, i.e. the exemplars, input text, and word meanings. We explicitly ask `GPT-5.1` to first decipher the input text and then translate it to English. We also provide it with the necessary steps in order to provide maximum aid. That is, we ask it to first guess the underlying language, then the method of ciphering, then decipher the input text, and finally, translate it to English.

We do this over a sample of 150 inputs, taken uniformly over 5 languages covering families and scripts (Hindi, French, Turkish, Telugu, Polish). We tried a few prompting settings, such wrapping the `LELemMS` prompt with additional deciphering instructions, as well as reframing the prompt entirely. These result in equivalent conclusions.

**Evaluation** We are interested in the following:

- Capability of the model to guess the correct language and method of ciphering: we manually check its guesses and cipher method reasoning.

- Capability of the model to uncover the ciphered text into the original source text: we evaluate BLEU score (Papineni et al., 2002) against the original input text.

- Translation quality of the final English translation: we calculate `chrF` score of the output translation against the reference, and compare it with `topline` (unciphered language performance). If the model indeed succeeds in undoing the cipher, its performance should be close to that of `topline`.

| Name | CL Name | Text | CL Text |
|---|---|---|---|
| Spanish | Serra | Esta es una frase de ejemplo. | Okjí ok áfí plíko yo osoñztó. |
| French | Gerestal | Ceci est une phrase d'exemple. | Qâqi âlz ùsâ nyfélâ h'âjâxnrâ. |
| Hindi | Manthi | यह एक उदाहरण वाक्य है। | मर अष ऊफुरकश डुष्म रे। |
| Czech | Krajčar | Toto je příklad věty. | Šášá fý qwósxím húšú. |
| Polish | Lukonik | To jest przykładowe zdanie. | Bu łąhb kjñisdonucą ñnoxęą. |
| Telugu | Nalaprika | ఇది ఒక ఉదాహరణ వాక్యం. | ఇప్టి అహా ఒఫేఇఔకఇ యేహ్ఇం. |
| Marathi | Sakpuri | हे एक उदाहरण वाक्य आहे. | री अष ऊफुरकश डुष्म इरी. |
| German | Markin | Dies ist ein Beispielsatz. | Nöuv övb uöm Xuóvlöupvobf. |
| Vietnamese | Noluа | Đây là một câu ví dụ. | Hēó tô dịl bēê jû gā. |
| Turkish | Sekman | Bu bir örnek cümledir. | Ta töb übvor qujkohöb. |

Table 2: Texts in ciphered languages (CLs).

**Results** We find that the model clearly fails in all regards.

- **Guessing language and cipher method:** For Hindi, Telugu, Turkish, and French, the model consistently guesses English as the underlying language. For Polish, it often guesses Polish as the underlying language. When the script is not Latin, it claims that the cipher method is a mapping of English words or characters on to the non-Latin script. It often also makes random guesses (Nepali /French / German for Hindi).

- **Deciphering input text:** BLEU scores between the deciphered input ciphertext and the original source text are close to random on the input text ($< 5$ BLEU for all languages), regardless of whether the language was guessed correctly. The model often attempts to back-translate provided word meanings into the guessed language rather than trying to decipher the ciphertext. When word meanings are left out, the hypothesis deciphered input is entirely random and unrelated to the source text.

- **English translation:** Accordingly, English translation scores are also very low. Provided all materials, they are comparable or the same as those of `LELemMS` performance (mean 36 `chrF` versus 42 for `LELemMS`, and 61 for `topline`), essentially obtained by relying on the word meanings and exemplars rather than decoding the ciphertext. When helper materials are not provided, produced translations consist of random unrelated text, similar to our `only-input` no-information baseline (23 mean `chrF` versus 26 for `only-input`, and $< 5$ BLEU score for all languages for both).

These results show that LLMs are far from capable of undoing an unknown monoalphabetic substitution cipher for an unknown source language, even with explicit instructions to do so. This is in line with the fact that decipherment of this type requires cryptanalytic expertise, and sufficient volume of data. It is possible that if trained to do so, prompted with more specific instructions, with enough data, and in the right inference setup, LLMs can perform such expert studies. However, when this is not the case, such as in our prompt setup where the model is explicitly presented with a scenario of a new natural language, with only a few exemplars, our experiments show that it is vanishingly unlikely that the LLM is able to perform any decipherment under the hood. Any observed performance, therefore, is due to language learning of the presented language.

**Observed behaviour** Manual review of outputs for all approaches indeed confirms that the LLM engages non-adversarially with our task setup. This is evident from our manual analysis of the qualitative improvements made in subsequent approaches, discussed in [Appendix H](#), along with examples for all approaches.

## B  Further experimental details

**Data**   We use a domain-balanced random subset of 100 samples per language for MT, and 200 samples per language for the other three tasks, as per our budget given costs of closed-source model inference and evaluation over all approaches, directions, and languages. These are fixed across across experiments.

**Evaluation**   We use the `GEMBA-MQM` implementation from (Kocmi and Federmann, 2023).[13] We use the `Unbabel/XCOMET-XL`[14] model for xCOMET, from Guerreiro et al. (2024). We use the `sacrebleu` (Post, 2018) implementation of `chrF`.

**Compute costs**   Our total costs associated with `OpenAI` API usage for inference and evaluation were $711, including exploration and final results.

We used `NVIDIA L40` GPUs for inference on `Llama-3.1-8B-Instruct` and `Qwen-2.5-7B-Instruct`. Presented results include 10 languages x 10 approaches x 2 directions = 200 experiments. Each experiment runs in approximately 15 minutes, totaling 50 GPU hours.

---

[13]`https://github.com/MicrosoftTranslator/GEMBA/`
[14]`https://huggingface.co/Unbabel/XCOMET-XL`

## C Details of all language learning strategies

### C.1 Additional approaches

We tried three other approaches with negative results, described here.

**Providing word sense**  We explore annotating target language words with word senses during X→English Machine Translation, to provide more context beyond lexical information. In practice, we retrieve multilingual WordNet definitions for each word in the input using the wn package (Goodman and Bond, 2021). However, the definition coverage of words proved to be too sparse for this approach to be effective.

**Syntax guidance via related language example**  In addition to $CL^{cov}$ELemMS, which uses a related high-resource language as a pivot, we also try a minimalistic approach, i.e. simply providing the related language translation as part of the context and asking the LLM to be guided by it for word order. This is potentially cheaper than $CL^{cov}$ELemMS since it does not require lexicons between the RHRL and the target language (alternatively, in the case that the latter are easier to obtain than lexicons with English, $CL^{cov}$ELemMS might be easier.) This approach is tested against $L^{cov}$ELemMS. This approach offers markedly less reduced benefits than $CL^{cov}$ELemMS. The LLM often fails to follow related language syntax closely and defaults to producing an English-influenced word ordering.

**Cheap reinflection**  We explore a cheap strategy for targeting verbal inflection for generation by additionally providing inflection paradigms to the model (LELemMSI). This information is much simpler to obtain than building morphological reinflection tools for a new language; and it is therefore worth testing how far it takes us. We test this for Hindi, German, Turkish, French and Spanish. See the inflection information provided in Table 4. Manual comparison of outputs with LELemMS shows several cases of better verb inflection with approach, although inconsistently and with several remaining problems, further discussed with examples in Appendix H. However, this effect is not strong enough for quantitative metrics to show gains in overall quality, and outputs continue to show several remaining problems in this regard due to inconsistent application of rules, as well as incomplete specification of potentially complicated verbal paradigms.

We highlight the significant scope for improvement in this regard.

### C.2 Materials

We describe the materials used for our approaches in more detail here.

**Lexicons**  We use ChiKhaPo lexicons[15] as described in § 5. For the Google Translate "lexicon" used in approach $L^{cov}$ELemMS and $CL^{cov}$ELemMS, we pass single words (without any context) into Google Translate API[16] and use the returned string as its lexicon equivalent. This simulates a lexicon with near-perfect coverage.

**Exemplars**  Since WMT24++ does not provide splits, we create our own split of of 300 samples, disjoint from the test split, fixed across languages and across approaches, as a pool for selecting exemplars at inference time. We note that this is the best case scenario, where the exemplar pool comes from the same distribution as the test set. In practice, such a resource may be difficult to construct.

**Named-entity recognition, lemmatizer and word features**  We use Stanza (Qi et al., 2020) on the languages inputs for NER, lemmatization, and morphological tags per word. Note that this is naturally run before ciphering. Any outputs of these tools consisting of language text (like lemmas) that are presented to the model are also ciphered and consistent with the ciphered language input.

**Syntax descriptions**  We provide an example of such a description in Table 3. Note that the explanation uses the (invented) ciphered language name when required, and is high-level enough that it does not distinguish the language from potentially other related languages in the family.

**Inflection paradigms**  We manually construct verbal inflection paradigms for LELemMSI covering agreement for 3 simple tenses and number, gender, and person, when relevant. We include running examples in the explanation itself. We include an example in Table 4.

---

```
Nalaprika is a Dravidian language.

Sentence-level word order (SOV): Nalaprika is typically SOV.

Object-verb order: (OV) The object precedes the verb.

Order of Adposition and Noun Phrase (Noun-Postposition): This language uses postpositions placed
after noun phrases.

Order of Genitive and Noun (Genitive-Noun): The genitive typically precedes the noun.

Order of Adjective and Noun (Adjective-Noun): Adjectives typically precede the noun.

Order of Relative Clause and Noun (Relative clause-Noun): Relative clauses typically precede the
noun.

Interrogatives: Polar questions do not use a dedicated question particle in the typology coding;
content questions use interrogative phrases without obligatory inversion.

Order of negation and verb (V-Neg): Negation typically follows the verb stem (i.e., is expressed
after V).

Morphosyntactic profile: Nalaprika is strongly suffixing and agglutinative, with productive
nominal case marking and rich verbal morphology.

Verbal inflection profile: Verbs inflect for tense/aspect/mood and agreement; complex predicate
structures are common.

Other characteristics: Case marking supports constituent reordering for information structure;
agreement is primarily expressed in the verbal domain rather than via gendered noun-adjective
agreement.
```

Table 3: High-level syntax overview of Telugu (ciphered name `Nalaprika`).

Table 4: Verbal inflection paradigms for French (ciphered name `Gerestal`) for LELemMSI. Note that that all French text, enclosed in "<>" is ciphered.

---

Inflection paradigm information for LELemMSI

---

```
Infinitival suffix: "<er>", "<ir>", "<re>"
Running examples (regular): "<parler>" (<er>), "<finir>" (<ir>/<re>-type patterns)

Present:
(<er> verbs)
- "<e>" (1st person singular) example: "<parle>"
- "<es>" (2nd person singular) example: "<parles>"
- "<e>" (3rd person singular) example: "<parle>"
- "<ons>" (1st person plural) example: "<parlons>"
- "<ez>" (2nd person plural) example: "<parlez>"
- "<ent>" (3rd person plural) example: "<parlent>"

(<ir>/<re> verbs; common regular pattern)
- "<is>" (1st person singular) example: "<finis>"
- "<is>" (2nd person singular) example: "<finis>"
- "<it>" (3rd person singular) example: "<finit>"
- "<issons>" (1st person plural) example: "<finissons>"
- "<issez>" (2nd person plural) example: "<finissez>"
- "<issent>" (3rd person plural) example: "<finissent>"

Simple Past:
(Formed with auxiliary "<avoir>" or "<être>" in the present + past participle.)
You must conjugate the auxiliary according to the subject, then add the correct past participle
form of the verb.

Auxiliary "<avoir>" (present forms):
```

---

- "<ai>" (1st person singular)
- "<as>" (2nd person singular)
- "<a>" (3rd person singular)
- "<avons>" (1st person plural)
- "<avez>" (2nd person plural)
- "<ont>" (3rd person plural)

Past participle endings:
- "<é>" (<er> verbs) example: "<parlé>"
- "<i>" (<ir> verbs) example: "<fini>"
- "<u>" (<re> verbs)

Full examples:
- "<ai parlé>" (1st person singular)
- "<as parlé>" (2nd person singular)
- "<a parlé>" (3rd person singular)
- "<ai fini>" (1st person singular)
- "<avons fini>" (1st person plural)
- "<ont fini>" (3rd person plural)

Simple Future:
(Formed by adding endings to the infinitive.)
- "<ai>" (1st person singular) examples: "<parlerai>", "<finirai>"
- "<as>" (2nd person singular) examples: "<parleras>", "<finiras>"
- "<a>" (3rd person singular) examples: "<parlera>", "<finira>"
- "<ons>" (1st person plural) examples: "<parlerons>", "<finirons>"
- "<ez>" (2nd person plural) examples: "<parlerez>", "<finirez>"
- "<ont>" (3rd person plural) examples: "<parleront>", "<finiront>"

### C.3 Constructing prompt

We provide the prompt templates for all our approaches for machine translation in Table 5. We also provide some additional notes here:

**Named entities** As mentioned in § 5, it is important that the model have access to named entities, since these are often trivially recognizable even in unseen languages. We identify named entities with Stanza. These are then transliterated into the output language script when the input and output language scripts are different. We include a glossary of named entities in the word meaning list regardless of approach.

**Translating into pivot RHRL for CL$^{cov}$ELemMS** We find that Google Translate gives high-quality translations from English to French or Spanish. However, for translating into Marathi, Hindi, Polish, or Czech, we use GPT-5.1 instead with a simple machine translation prompt.

See Appendix I for prompts for other tasks.

Table 5: Example prompts for English and Spanish, ciphered language name `Serra`. We show the (1) topline (`topline`) that simply uses Spanish. We show (2) LELemMS for →eng and (3) LELemMS for eng→as approaches with all used materials. Other approaches use the relevant subset of materials. We also show (4) $CL^{cov}$ELemMS for eng→ , demonstrating the cascade through a high-resource related language (French in this case).

---

`Approach prompts`

---

(1) `topline`: →`eng` (Spanish → English)

---

```
Translate the following text from Spanish to English. Respond only with the translation.
Input:
Mi hermano vino y desmanteló todo, y estuve postergando empezar a trabajar juntos para rehacerlo
de cero.

Output:
```

---

(2) `LELemMS`: →`eng` (Spanish → English)

---

```
Serra is a newly discovered Romance language.

Translate the following text from Serra to English. Respond only with the translation.
Input:
Ñi golñífó difó q yokñífjota jóyó, q okjádo zókjolvífyó oñzowíl í jlíhísíl sáfjók zílí logíxoltó
yo xoló.

Here are some examples of inputs and outputs:
Input1: Tík xoldoxolík ko ...
Output2: Brewers froth over ...
Input2: Of tí ójlí ...
Output2: The other one ...
Input3: Ñék íyotífjo ...
Output3: Up ahead ...
...

Here are some word meanings for words in the input text. Note that these may not be correctly
inflected. Also note that there may be multiple translations for a word, in which case you
should choose the most relevant one.
ñi - my,e
golñífó - brother,sibling
difó - came,wine
dofil - come,arrive
...

Here is POS, lemma, and grammatical feature information for words in the input text:
Ñi: POS: DET, Lemma: ñi, Features: Number=Sing|Person=1|Poss=Yes
golñífó: POS: NOUN, Lemma: golñífó, Features: Gender=Masc|Number=Sing
difó: POS: VERB, Lemma: dofil, Features: Mood=Ind|Tense=Past|Person=3
yokñífjota: POS: VERB, Lemma: yokñífjotíl, Features: Mood=Ind|Tense=Past|Person=3
...

Here is a general description of the syntax of the language Serra:
Serra is a Romance language.
Sentence-level word order (SVO): Serra is typically SVO.
Object-verb order: (VO) The verb precedes the object.
...
```

---

---

Consider all the information provided above. Respond with *only* the output.
Input:
Ñi golñífó difó q yokñífjota jóyó, q okjádo zókjolvífyó oñzowíl í jlíhísíl sáfjók zílí logíxoltó
yo xoló.
Output:

---

**(3) LELemMS: eng→ (English → Spanish)**

Serra is a newly discovered Romance language.

Translate the following text from English to Serra. Respond only with the translation.
Input:
My brother came and ripped it all out and I've been putting off starting to work to put it all
back together

Here are some examples of inputs and outputs:
Input1: Time for the ...
Output1: Ttova ot ñóñofjó ...
Input2: My surgeon is ...
Output2: Ñi xilásífó ok ñáq ...
Input3: Up ahead ...
Output3: Ñék íyotífjo ...
...

Here are some word meanings for words in the input text. Note that these may not be correctly
inflected. Also note that there may be multiple translations for a word, in which case you
should choose the most relevant one.
my - ñi,qó
brother - golñífó,íxóñzírífjo
came - difó,dofil
...

Here is POS, lemma, and grammatical feature information for words in the input text:
My: POS: PRON, Lemma: my, Features: Case=Gen|Number=Sing|Person=1|Poss=Yes
brother: POS: NOUN, Lemma: brother, Features: Number=Sing
came: POS: VERB, Lemma: come, Features: Mood=Ind|Tense=Past|Person=3
and: POS: CCONJ, Lemma: and, Features: -
...

Here is a general description of the syntax of the language Serra:
Serra is a Romance language.
Sentence-level word order (SVO): Serra is typically SVO.
Object-verb order: (VO) The verb precedes the object.
...

Consider all the information provided above. Respond with *only* the output.
Input:
My brother came and ripped it all out and I've been putting off starting to work to put it all
back together
Output:

---

**(4) CL$^{cov}$ELemMS: eng→ (English → French → Spanish)**

---

Serra is a newly discovered Romance language.

Translate the following text from French to Serra. Respond only with the translation.
Input:
Mon frère est venu et a tout arraché et j'ai retardé le début du travail pour tout remonter.

Here are some examples of inputs and outputs:
Input1: C'est l'heure de ...
Output1: Ttova ot ñóñofjó ...
Input2: Mon chirurgien est ...
Output2: Ñi xilásífó ok ñáq ...
Input3: Devant, ...
Output3: Ñék íyotífjo, ...
...

Here are some word meanings for words in the input text. Note that these may not be correctly inflected. Also note that there may be multiple translations for a word, in which case you should choose the most relevant one.
mon - ñi
frère - golñífó
est - okjo
venu - difó
...

...

Consider all the information provided above. Respond with *only* the output.
Input:
Mon frère est venu et a tout arraché et j'ai retardé le début du travail pour tout remonter.
Output:

## D Scores for other models

We show `chrF` and `xCOMET` performance for `Llama-3.1-8B-Instruct` and `Qwen-2.5-7B-Instruct` in Figure 5. We see that the improvements for →eng for these smaller models are much lower, indicating that they may not have the capabilities to take advantage of linguistically complex instructions.

See our GitHub repository for all outputs and associated scores.

## E Domain experiments

Our framework allows us to gain more fine-grained insight onto performance of ICLL strategies by letting us avail of rich frontier benchmarks. This lets us explore, for example, whether there are domain disparities in performance, using domain diversity in `WMT24++`. We provide a breakdown of `xCOMET` scores by domain in Figure 6. This shows a clear imbalance by domain, with the model performing better on `social` and `literary` than others, and particularly badly on news or speech-like text. Insights of this type may indicate gaps in materials (lexicons, exemplars) or approaches that can direct future strategies or resource development depending on the use case.

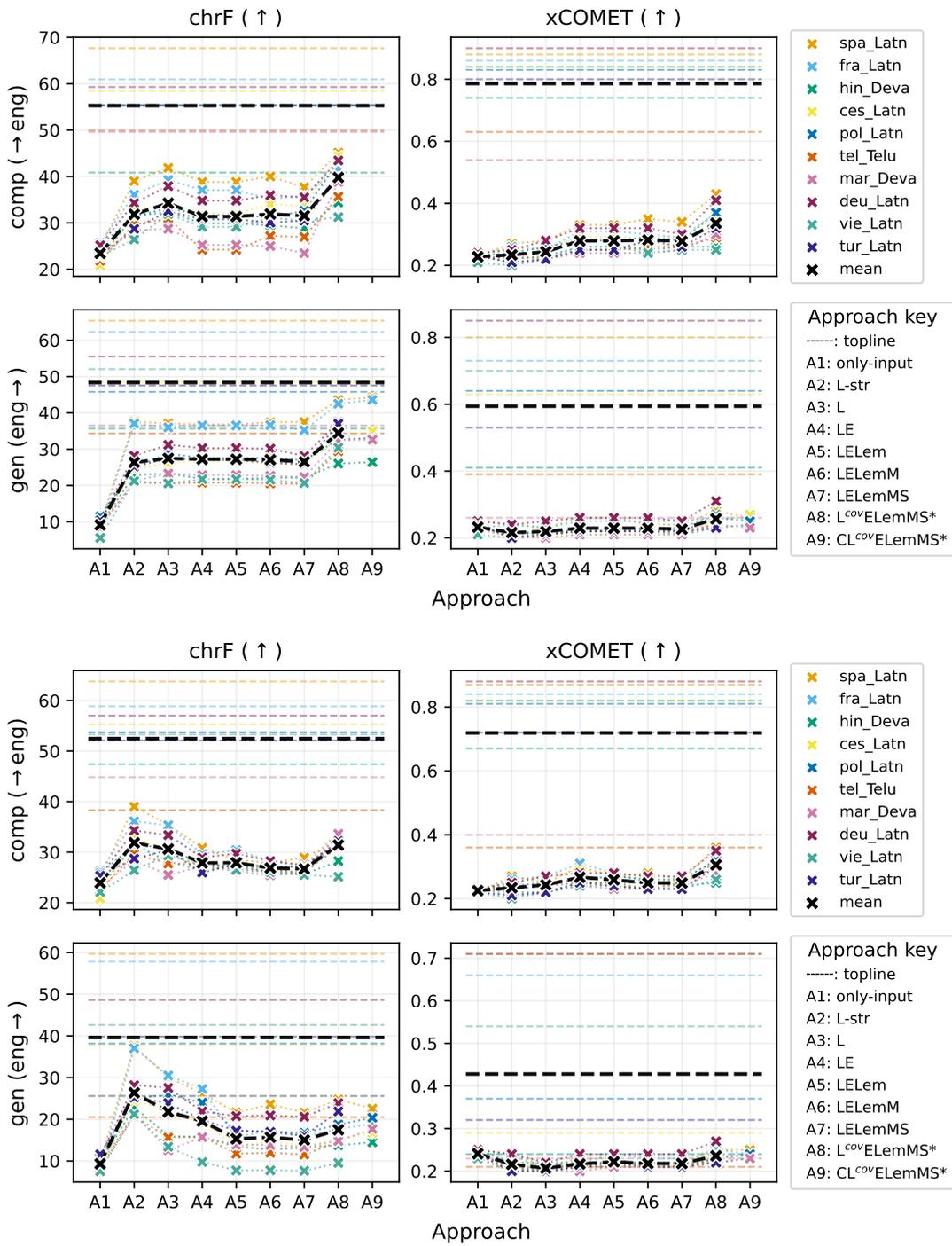

Figure 5: Results for `Llama-3.1-8B-Instruct` (top) and `Qwen-2.5-7B-Instruct` (bottom).

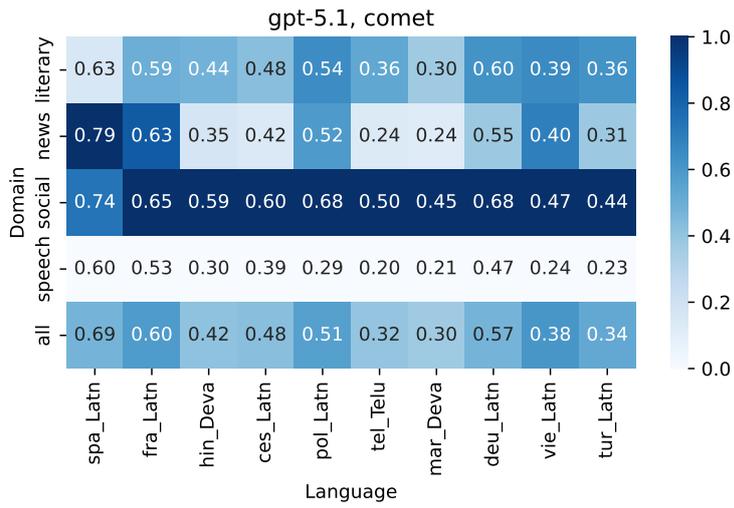

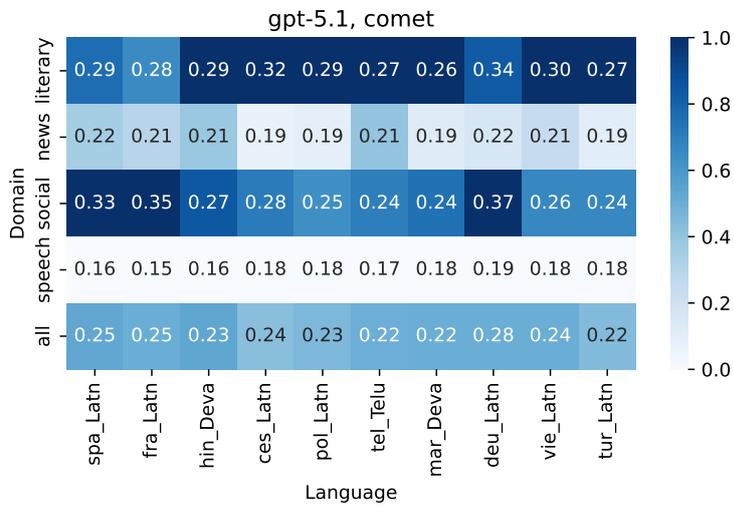

Figure 6: Performance across domains for →`eng`(top) and `eng`→(bottom). Darker blue indicate better performance, normalized by language.

## F Calibrating `GEMBA-MQM` judgments

`GEMBA-MQM` is an LLM-based evaluation metric and is not guaranteed to be reliable, especially for all our languages on the generation side. While it is out-of-scope for us to provide meta-evaluation at scale for `GEMBA-MQM` for our languages, we perform a small analysis to sanity check the general signal provided by this metric.

**For →eng** Firstly, we note that for →eng, the model is simply judging English outputs, given source texts in various languages. The model is quite capable of understanding source texts in these mid-to-high resource languages, and is consistent in its severity judgments for English outputs.

**For eng→** For eng→, where it has to judge outputs in various languages, `GEMBA-MQM` scores for all languages for the ciphered settings remain close to the worst possible score. This is very reasonable given their quality, as we judge in our manual analysis. We look to the unciphered (`topline`) setting to get a sense of the reliability of `GEMBA-MQM` for these languages. We go through `GEMBA-MQM` judgments for 30 outputs for both comprehension and generation for Marathi and French, which are one of the lowest and highest-resource languages in our set respectively. In general, we find that the LLM is sometimes excessively strict, counting small deviations from style or wording as minor errors. In this regard, we find that scores for Marathi and Hindi are somewhat lower than those for French and Spanish, simply because they have less literal translations and accumulate more minor "errors" from `GEMBA-MQM`, despite being of similar quality. However, the model does so consistently for a given language, and relative scores are therefore reasonably reliable.

Thus, we would caution against comparing scores across languages. However, we concur in the unusability judgments (i.e. critical errors, resulting in a score of 25) provided by `GEMBA-MQM` for most generation outputs, and believe that score differences from `GEMBA-MQM` for samples in our approaches generally correspond to concrete and noticeable quality differences between outputs.

We acknowledge that these are merely observations based on a very small sample set, and heavier reliance on LLM-based evaluation metrics for particular languages require at-scale meta-evaluations for those languages. This is a general limitation of LLM-as-judge metrics (Kocmi and Federmann, 2023).

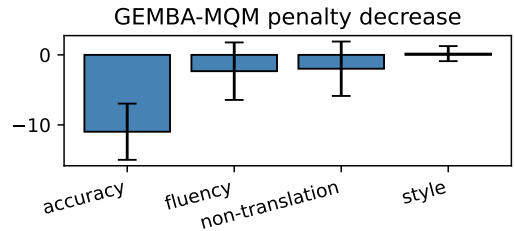

Figure 7: Improvements in various `GEMBA-MQM` (↓) categories for LELemMS compared to L.

## G Analysis: What characteristics of outputs improve?

`GEMBA-MQM` provides an explanation and categorization of error types ("accuracy", "fluency", which have further subdivisions) and severity levels ("critical", "major", "minor"), providing scores for interpretable categories such as accuracy and fluency. This allows us to investigate the nature of improvements we see in different approaches. We looked at the non-tied subset of samples as per `GEMBA-MQM` for LELemMS vs the baseline L approach, and calculate points per category, capped at 25.

**For →eng** We find that score improvements for comprehension come almost entirely from accuracy points. See Figure 7. This makes sense, since LLM outputs in English are fluent even for baseline approaches with very low adequacy. However, there are still slight grammar and stylistic improvements that are not explained by lexical choice, when the LLM understands the input better, presumably due to lowered confusion from seemingly garbled or incomprehensible input.

**For eng→** There are very few, typically less than 5% of outputs for eng→that are not tied at 25 (lowest score), regardless of approach compared. We compute uncapped scores for error categories, and find that on average each sample is over 25 for both accuracy and fluency, for all approaches, i.e. outputs are critically flawed in terms of both meaning and grammaticality. Justifications provided by `GEMBA-MQM` typically complain about complete incomprehensibility and nonsensicality of the output, validated as reasonable by our analysis in Appendix F.

# H  Manual analysis

We include our manual review of samples here as mentioned in § 6.

**Goal**    The goal of the analysis is to characterize the qualitative nature of the improvements made by each approach (if any), including all current and new strategies tried in the paper, and to present examples.

**Description of analysis**    Each of our approaches adds a resource or strategy incrementally to a previous approach. We therefore compare outputs pairwise between two approaches differing only in the incremental addition in order to understand the benefits of the addition. For example, we compare LELemM against LELem, and LELemMS against LELemM. We compare LELemMSI and $L^{cov}$ELemMS against LELemMS, since they add inflection paradigm descriptions and a lexicon change to LELemMS, and we compare $L^{cov}$ELemMS to $CL^{cov}$ELemMS. We do so over four languages: Hindi, Marathi, French, and Turkish. The analysis was performed by the authors, who are native speakers of or fluent in the studied languages.

**Qualitative findings**    We obtain a qualitative sense of the effects of each approach, and provide our findings below. We provides examples accompanying each of these points in Table 6. **These examples are cherry-picked** to illustrate the benefits of each approach, and we refer the reader to quantitative metrics instead to understand the overall extent of translation quality improvement.

- Baselines: Table 6 also contains representative (non-cherry-picked) examples of outputs from topline and baseline approaches, showing that `topline` is fluent and accurate, that `only-input` produces random unrelated text, and word-for-word translation with `L-str` is nonsensical and invalid.

- LE: As expected, lemmatizers help to increase coverage of the input words, when the lexicon may not cover an inflection but does have the lemma.

- LELem: Exemplars, chosen with BM25 string similarity, generally help with particular overlap strings in the input. They may contain a particular expression, or the usage of some function words correctly, that the model copies over and therefore improves over literal translation. Occasionally, the exemplars also have more intangible effects, such as helping the model not deviate from the subject, if they share the same subject as the input. In general, in-domain exemplars help considerably with performance given overlap with the input in terms of topic and surface similarity. Note that this makes the exemplars quite an expensive resource to collect, i.e. bitext covering a diverse range of subjects and vocabulary.

- LELemM: Added word features and POS tags help in targeted scenarios where the word meaning in the lexicon is ambiguous by part-of-speech. The model uses the tense or other information occasionally, but not consistently.

- LELemMS: In the context of generation - the model uses the high-level syntactic information to reorder certain phrases, especially for languages with very different word from English (like Hindi and Marathi). This is especially true in simple phrases where, for example, there is only a very simple subject, object, and verb, or a noun and adjective. As sentences get longer, however, this information is no longer enough, and the model very quickly fails to apply the guidance even in simple phrases within longer inputs. In general, word order remains very much mangled, even with some attempts to reorder as per the guidance. For comprehension, this information helps the model to correctly interpret events in certain cases by understanding, for example, subject and object ordering.

- LELemMSI: The model attempts to inflect verbs as per the provided guidance. However, there remain several obstacles to perfect verb inflection. Firstly, given that the model often fails to correct interpret concepts and events, it often does not have the information (about subject, object, gender, number, aspect, etc.) that it needs to perform correct verb agreement, even if it knows the correct inflection in every case. Secondly, the model displays inconsistent success in drawing from the guidance, and fails to do so as inputs gets longer.

- $L^{cov}$ELemMS: Google Translate lexicons naturally help considerably and straightforwardly

in providing coverage over missing words and morphological variants.

- $\texttt{CL}^{cov}\texttt{ELemMS}$: The related language cascade has perhaps the most success in mitigating many of the above problems. The related language in question often has similar idiomaticity and lexical usage patterns, that may differ from those of English. While English phrasing often results in nonsensical literal translations into Hindi, a fluent English-Marathi translation transfers it into phrasing that is much likelier to result in much better Hindi output when literally translated into Hindi. This is also true for morphological paradigms: these often map much better between related languages. Finally, this approach also naturally handles the word-order issue with much more success. Having said this, there are some important caveats here. Firstly, in the case where the target ciphered language is already somewhat close to English, the cascade may hurt by introducing noise and error propagation. Secondly, literal translation even between related languages is very noisy and of poor quality, and heavily depends on the extent of their relatedness. We observe that Hindi and Marathi are much more helped by this approach than French in all of the above aspects.

In general, we find that the benefits of each approach are by no means consistent and systematic over all inputs or even within a single input, and there is large scope of improvement beyond these.

Table 6: Examples of MT outputs per approach, demonstrating performance of baselines (`topline`, `only-input`, `L-str`), as well as qualitative benefits of each incrementally added material.

| Approach | Dir | Sample | Comment |
|---|---|---|---|
| `topline` | →eng | **Input:** Mein Bruder war da und hat alles auseinandergenommen, und ich schiebe immer weiter auf, das alles wieder zusammenzubauen<br>**Reference:** My brother came and ripped it all out and I've been putting off starting to work to put it all back together<br>`topline:` My brother was here and took everything apart, and I keep putting off reassembling it all. | Fluent and accuracy translation. |
| `topline` | eng→ | **Input:** And I have a lot of experience. I've worked internationally. I've worked this continent. I've worked different states, different market segments. And what I found is people are really intrigued by that. They want to know kind of what the behind the scenes is. So some of these videos are going to be tips from somebody that actually works in this industry, and I think that's what separates me a little bit from possibly other travel bloggers that you might follow. You know, I may not have worked for, say, even the cruise industry, but I had to take courses in it for my hospitality degree. So, even if I haven't worked in it, I've been educated on it.<br>**Reference:** J'ai beaucoup d'expérience. J'ai travaillé à l'international, j'ai travaillé sur ce continent, j'ai travaillé dans différents États et différents segments de marché. Et ce que j'ai découvert, c'est que cela intrigue beaucoup les gens. Ils veulent savoir ce qui se passe en coulisse. Certaines de ces vidéos seront des conseils de la part de quelqu'un qui a l'avantage de travailler dans ce secteur, et je pense que c'est ce qui me distingue des autres blogueurs de voyage que vous suivez peut-être. Vous savez, je n'ai peut-être pas travaillé dans le secteur des croisières, par exemple, mais j'ai suivi des cours dans ce domaine quand je préparais mon diplôme d'hôtellerie. Alors, même si je n'ai pas travaillé dans ce domaine, j'ai les connaissances.<br>`topline:` Et j'ai beaucoup d'expérience. J'ai travaillé à l'international. J'ai travaillé sur ce continent. J'ai travaillé dans différents États, différents segments de marché. Et ce que j'ai constaté, c'est que les gens sont vraiment intrigués par ça. Ils veulent un peu savoir ce qui se passe en coulisses. Donc, certaines de ces vidéos seront des conseils venant de quelqu'un qui travaille réellement dans cette industrie, et je pense que c'est ce qui me distingue un peu peut-être d'autres blogueurs de voyage que vous suivez. Vous savez, je n'ai peut-être pas travaillé, disons, dans l'industrie des croisières, mais j'ai dû suivre des cours à ce sujet pour mon diplôme en hôtellerie. Donc, même si je n'y ai pas travaillé, j'ai été formé(e) dessus. | Fluent and accurate translation. |
| `only-input` | →eng | **Input:** Mon frère a tout arraché, et j'ai laissé traîner avant de m'y mettre<br>**Reference:** My brother came and ripped it all out and I've been putting off starting to work to put it all back together<br>`only-input:` This house is very small and dark, but it feels pleasantly warm in the winter | Zero-information baseline, output is random and unrelated to input, but following length and punctuation cues. |
| `only-input` | →eng | **Input:** त्यासाठी लायब्ररी किंवा अजून काही लागणारसुद्धा नाही.<br>**Reference:** don't even need a library for it or anything<br>`only-input:` The wise river carries stories to the distant mountains. | Zero-information baseline, output is random and unrelated to input, but following length and punctuation cues. |
| `L-str` | →eng | **Input:** चलिए, बातचीत को थोड़ा पीछे ले चलते हैं. यह किसी ने भी नहीं पूछा, "इसमें वास्तव में है क्या?".<br>**Reference:** But let's rewind a bit. "What's actually in there?" I hear nobody ask.<br>`L-str:` sake conversation to little behind carry stay are it certain by also no ask therein really in is what splutter | Non-LLM word-for-word translation baseline fails. |
| `L-str` | eng→ | **Input:** But let's rewind a bit. "What's actually in there?" I hear nobody ask.<br>**Reference:** Mais, petit flash-back : "Qu'est-ce qu'il y a dedans ?", n'entends-je pas demander.<br>`L-str:` mais laisser rembobinage un bit quoi en fait dans je entendre personne demander | Non-LLM word-for-word translation baseline fails. |

*Continued on next page*

| Approach | Direc-tion | Sample | Comment |
|---|---|---|---|
| LE | →eng | **Input:** @user12 @user11 सबको साथ लेकर चलना और सबकी सुरक्षा का ध्यान रखना, ये हमारी प्राथमिकता होनी चाहिए. आप किस प्राथमिकता" की बात कर रहे हैं? <br>**Reference:** @user12 @user11 inclusion and safety are always a good thing to prioritize. what default you are referring to? <br>**L:** @user12 @user11 Proceed with the text with accompaniment, and keep the lesson's safety in attention; our priority ought to be so. What do you mean by "priority"? <br>**LE:** @user12 @user11 lessons should proceed with <span style="color:green">inclusion and keep safety in mind</span>; these ought to be our priority. So what are we doing when we say "priority"? | Exemplars include the words "inclusion" and "safety". |
| LE | →eng | **Input:** लेकिन आप यह लज़ीज़ ब्रेड खाने के लिए, अपने दोस्तों या रिश्तेदारों को घर पर बुला सकती हैं ना? <br>**Reference:** surely, though, you can invite people round to help you devour that beautiful work of bread-y art, no? <br>**L:** But if you miss this delicious bread for your own sake, will <span style="color:red">your own friends or relatives summon you to their homes</span> for cotton? <br>**LE:** But if you're not eating this delicious bread yourself, are <span style="color:green">you inviting friends or relatives over</span> for it? | Examples include food-related text. LE is able to stick to the topic and avoid being misled by noise from lexicon ("cotton") |
| LELem | →eng | **Input:** Jetzt überlege ich, ob eine Million Meter in 2024 drin sein könnten? #peloton <br>**Reference:** Now I wonder if a million meters in 2024 might be possible? #peloton <br>**LE:** Right now I'm wondering: could there be a million meters in 2024? #peloton <br>**LELem:** Now I'm wondering if a million meters in 2024 <span style="color:green">might be possible</span>? #peloton | Lemma coverage allows the correct semantic idea to be conveyed. |
| LELem | →eng | **Input:** Der Zahnarzt war nicht zufrieden damit, wie die Krone sitzt, selbst nachdem er sie bearbeitet und gefeilt hatte. Deshalb haben sie neue Abdrücke gemacht und es wird eine neue Krone angefertigt. <br>**Reference:** Dentist wasn't happy with the fit of the crown, even after he'd footered and filed it. So new impressions were taken, and a new crown will be made. <br>**LE:** The dentist wasn't happy that the normal crown sat badly, because he had <span style="color:red">machined it</span> himself in a rush. Therefore you have new imprints made, and he gets a new artificial crown. <br>**LELem:** The dentist wasn't happy that the normal crown sat badly, because he himself had hurriedly <span style="color:green">filed it</span>. So <span style="color:green">he had a new imprint made</span>, and he got a newly fabricated artificial crown. | |
| LELem | →eng | **Input:** Oh schaut! Das Ding ging vor ungefähr einer Stunde, für etwa eine halbe Minute wieder online. <br>**Reference:** Hey, look: the thing went back online for like half a minute about one hour ago. <br>**LE:** By Jove! The thing came back online <span style="color:red">after roughly an hour</span>, maybe about <span style="color:red">half a minute earlier</span>. <br>**LELem:** By Jove! The thing came back on <span style="color:green">roughly an hour ago</span>, and has been online again for about half a minute. | Verb lemmas in LELem help to get the meaning across. |
| LELem | eng→ | **Input:** Last week the weather didn't cooperate so we're back at it today. <br>**Reference:** Letzte Woche hat das Wetter nicht mitgespielt, also versuchen wir es heute noch mal. <br>**LE:** Last week aber wetter <span style="color:red">nuss</span> tatenbeistehen also wir antwort rücken a das today. <br>**LELem:** Last week aber wetter taten <span style="color:green">kein</span> beinflussen auch wir antwort rücken à das today, | Lemmatization allows for explicit listing of the negative, which shows up in LELem (although not as the correct lexical choice here for the negative) but not LE. |

| Approach | Direc-tion | Sample | Comment |
|---|---|---|---|
| LELemM | →eng | **Input:** तिला गार्ड्स साइटकडे घेऊन जात असताना, गावातील लोक त्यांच्या एकमेव आणि आवडत्या राणीला मारण्यासाठी घेऊन जाताना पाहून खूप रडत होते. काही लोक रडत होते, काही लोक ओरडत होते आणि गार्ड्सवर हल्ला करण्याचा प्रयत्न करत होते आणि इतर आश्चर्यचकित होऊन पाहत होते की हे का घडत आहे; परंतु त्या सर्वांनी तोच मुद्दा धरून ठेवला होता.<br>**Reference:** As the guards escorted her to the site, the townspeople cried out in outrage at seeing their only queen- and the nicest one- about to be killed. Some people were crying, some people were shouting and trying to attack the guards, and others were wondering why this was happening; but they all shared the same basis.<br>**LELem:** Her guards were of a different caste, and the villagers' sole and favorite queen was being beaten with sixty warm rods, which caused much crying. Some people cried, some people shouted, and the guards who carried out the raid savored what they had done, and others were astonished by what they saw, so it was; but she knew all this only too well.<br>**LELemM:** Her guard had a habit of riding bicycles, but that day he had gone to the queen's favorite maid on his own and they were very upset. Some people were upset, some people were shouting, and they said the guard leading the raid was to blame and that something astonishing had happened, so that was exactly what took place; but they all only pretended to know it. | LELemM correctly shows progressive past actions ("were shouting") due to having access to morphological information. It is also able to maintain the correct subject ("they") due to grammatical information. Note that both outputs have significant hallucinations. |
| LELemM | →eng | **Input:** परंतु," तो पुढे म्हणाला. तुला फाशी दिली जाइल."<br>**Reference:** "But," he added. "You will be executed."<br>**LELem:** "But," he said before us. "You will go to the execution in Chile."<br>**LELemM:** "But," he went on, "you will not be put to death." | LELemM reduces one hallucination due to having access to part-of-speech and word features of auxiliaries, but introduces a hallucinated negative, changing the meaning of the sentence. |
| LELemM | →eng | **Input:** La semana pasada el tiempo no quería cooperar, así que hoy volvemos al tema.<br>**Reference:** Last week the weather didn't cooperate so we're back at it today.<br>**LELem:** This week went by without us really getting to work together, so we'll have to come back to that topic another day.<br>**LELemM:** The past week the weather wasn't really conducive to working together, so today we're coming back to the issue. | LELemM is able to maintain "weather" as the subject, and communicate main ideas due to this. |
| LELemM | →eng | **Input:** El dentista no estaba contento con el ajuste de la corona, incluso después de dedicarle un buen rato y haberla limado. Así que tomó nuevas muestras y hará una corona nueva.<br>**Reference:** Dentist wasn't happy with the fit of the crown, even after he'd footered and filed it. So new impressions were taken, and a new crown will be made.<br>**LELem:** the dentist wasn't very happy with her crown fitting, even after spending a good while on the filing. that's how I got the news sample and made another new crown.<br>**LELemM:** The dentist wasn't very happy with the adjustment of her crown, even after spending a good while filing it down. That's how he took new impressions and will make her another new crown. | POS of the word "nuevas" helps disambiguate the sense of the word and LELemM therefore has the correct adjective sense ("new impressions") while LELem has the wrong sense and mistranslates the phrase ("news sample"). |
| LELemMS | eng→ | **Input:** Wish me luck!<br>**Reference:** Bana şans dileyin!<br>**LELemM:** arzu beni alınyazısı<br>**LELemMS:** beni alınyazısı arzu | (Turkish is SOV) LELemMS shows more SOV-like tendency by placing the object-like element ("beni") before the predicate-like element ("arzu"). Still, the output remains nonsensical due wrong object case ("beni") and wrong word-sense ("arzu"). |

| Approach | Direction | Sample | Comment |
|---|---|---|---|
| LELemMS | eng→ | **Input:** "You mean... the horrible god monster thing that we... helped kill?" Tenuk stuttered.<br>**Reference:** "Tu veux dire... l'horrible monstre dieu que nous avons... aidé à tuer ?", bégaya Tenuk.<br>**LELemM:** "Tu moyenne... le <span style="color:red">horrible dieu</span> apparition chose ce me... aidé tuer ?", Tenuk bégayer.<br>**LELemMS:** "Tu moyenne... le chose apparition <span style="color:green">dieu horrible</span> ce me... aidé tuer ?", Tenuk bégayer. | LELemMS shows noun-adjective order, correct for French. |
| LELemMS | eng→ | **Input:** Might be the odd cheater in my family, but considering my largely broke folks' roots, any affairs were probably in mostly homogenous villages, small towns, and ethnic ghettos lol.<br>**Reference:** Je suis peut-être le seul membre de ma famille à être infidèle, mais si je m'en tiens aux racines de mes parents miséreux, les infidélités avaient probablement lieu dans des villages, des petites villes et des ghettos ethniques plutôt homogènes, MDR.<br>**LELemM:** Pourrait être le étrange charlatan dans mon famille, mais vu mon amplement fauché gens de aïeux, tout affaires étais apparemment dans <span style="color:red">généralement homogène village</span>, petit ville et barbare favelle mdr.<br>**LELemMS:** Pourrait être le atrange charlatan dans mon famille, mais vu mon amplement fauché gens légume, tout affaires étaient apparemment dans <span style="color:red">généralement homogène village</span>, petit ville et barbare favelle mdr. | With longer inputs, the model fails to reorder adjective noun phrases as noun-adjective despite explicit instruction. |
| LELemMS | →eng | **Input:** थसालीन ने नियसी को अपनी पूंछ से खींचकर पीठ पर बिठा लिया और वहां से उड़ चला. हमला करने वाला वह काला जीव भी थसालीन का पीछा करने लगा.<br>**Reference:** Thassalin responded to this attack by grabbing Nyssi and throwing her on his back with his insanely long, dextrous tail, before taking off into the air. The black, two-winged creature quickly chased after them.<br>**LELemM:** <span style="color:red">Thassalin let Nyssi grab hold</span> of his tail, then carried her along on his back. The attacking dark creature, however, only focused on pursuing Thassalin.<br>**LELemMS:** <span style="color:green">Thassalin had grabbed Nyssi</span> by her tail and lifted her up onto his back, then had taken off with her. The attack allowed the young dark creature to catch up with Thassalin. | The model is able to use high-level syntactic knowledge to identify "Thassalin" as the actor and "Nyssi" as the object. This makes for a generally better translation, though still incorrect. |
| L$^{cov}$ELemMS | →eng | **Input:** Az önce elektrik süpürgesini tekrar çalıştırdı. Umarım iyiye işarettir.<br>**Reference:** He just started vacuuming again. I hope it's a good sign.<br>**LELemMS:** At first there was barely any <span style="color:red">power to run the baseboard heater</span> again. Hopefully it'll get better soon.<br>**L$^{cov}$ELemMS:** A little while ago <span style="color:green">electric broom</span> worked again. <span style="color:green">I hope</span> it's a <span style="color:green">sign</span> of things getting better. | L$^{cov}$ELemMS conveys the topic relatively better ("electric broom" closer to "vaccuum") due to better lexicon coverage. It also removes the hallucinated content ("there was barely any power"). |
| L$^{cov}$ELemMS | →eng | **Input:** Bana şans dileyin!<br>**Reference:** Wish me luck!<br>**LELemMS:** To me, fortune, proclaim it!<br>**L$^{cov}$ELemMS:** <span style="color:green">Wish me luck!</span> | L$^{cov}$ELemMS 's expanded lexicon directly supports input words, allowing for the right interpretation. |
| LELemMSI | eng→ | **Input:** Politicians who think they can overturn judges' decisions are effectively putting themselves above the law<br>**Reference:** Les hommes politiques qui estiment pouvoir contourner les décisions de justice se placent de fait au-dessus des lois<br>**LELemMS:** Politicien ce qui <span style="color:red">penser ils</span> conserve abattre choix les juges sont activement mise eux-mêmes au-dessus des loi<br>**LELemMSI:** Politicien qui <span style="color:green">pensent</span> les conserve pout abattre choix des juges sont activement lancement eux au-dessus de lois | LELemMSI uses the correct third person plural verb "pensent" (to think), whereas LELemMS uses an uninflected/ill-formed "penser ils". |

| Approach | Direction | Sample | Comment |
|---|---|---|---|
| `LELemMSI` | eng→ | **Input:** We're just going to let that sit and rehydrate for a couple minutes. With ultralight cooking, we're not actually cooking on the stove. We're primarily boiling water so that we can rehydrate our food. It's also one of the reasons why we use instant grits. Less cooking time, less fuel, and less weight. Oh yeah. So this is about what we've got going on here. Perfect consistency.<br>**Reference:** Nous allons laisser ça reposer et se réhydrater pendant quelques minutes. Dans la cuisine ultra légère, on ne cuisine pas vraiment sur le réchaud. On fait surtout bouillir l'eau qui va venir réhydrater notre nourriture. C'est aussi une des raisons pour lesquelles on utilise du gruau instantané. Moins de temps de cuisson, moins d'énergie et c'est moins lourd. Excellent ! Voilà ce que ça donne. La consistance est parfaite.<br>**LELemMS:** Nous sont juste train pour laisser ce s'asseoir et hydrater pour a couple minutes. Avec ultraléger cuisson, nous sont négation en fait cuisson sur le poêle. Nous sont avant bouillante eau donc que nous conserve hydrater nourriture notre. Ça de aussi un de accident pourquoi <span style="color:red">nous utiliser</span> instant gruau. Moins temps cuisson, moins alimenter, et moins pondérer. Oh, ouais. Donc ce tout à propos de quoi <span style="color:red">nous avoir</span> obtenu partie. Parfait consistent.<br>**LELemMSI:** Nous parlons juste aller pour laisser ce s'asseoir et hydrater pour a couple minutes. Avec ultraléger cuisson, nous pas crônat cuisson sur le poêle. Nous sont avant bouillante eau donc ce nous <span style="color:red">conserve</span> hydrater notre nourriture. Ça est aussi un de accident pourquoi <span style="color:green">nous utilisons</span> instant gruau. Moins temps cuisson, moins alimenter, et moins pondérer. Oh, ouais. Donc ce ça est à propos de ce que <span style="color:green">nous avons</span> posséder train ici. Parfait consistent. | LELemMSI uses the correct first person plural verb forms while LELemMS uses wrong inflections. However, note that LELemMSI also contains several badly inflected forms, showing imperfect rule-following, or imperfect rule coverage and precision. |
| `CL`cov`ELemMS` | eng→ | **Input:** Initial applications to Harvard for a psychology master's were rejected, but was eventually admitted. The initial setbacks were due to Milgram not taking any undergraduate courses in psychology at Queens College. In 1961, Milgram received a PhD in social psychology. He became an assistant professor at Yale around the same time.<br>**Reference:** हार्वर्ड विश्वविद्यालय में मनोविज्ञान से मास्टर्स के के लिए, मिल्ग्राम के शुरुआती आवेदन खारिज कर दिए गए थे. हालांकि, आखिरकार उन्हें दाखिला मिल गया था. मिल्ग्राम को शुरू में दिक्कतें इसलिए आईं, क्योंकि उन्होंने क्वींस कॉलेज में मनोविज्ञान में कोई अंडरग्रैजुएट कोर्स नहीं किया था. साल 1961 में मिल्ग्राम ने सामाजिक मनोविज्ञान में पीएडी की डिग्री हासिल की. इसी दौरान वह येल विश्वविद्यालय में असिस्टेंट प्रोफ़ेसर भी बन गए.<br>**L**cov`ELemMS:` प्रारंभिक अनुप्रयोग हार्वर्ड के लिए ए मनोविज्ञान मास्टर का <span style="color:red">में थे अस्वीकार कर दिया</span>, लेकिन अंततः स्वीकार किया. प्रारंभिक असफलताओं मिल्ग्राम में क्वींस कॉलेज पर मनोविज्ञान में कोई अवर पाठ्यक्रम <span style="color:red">ले रहा हूँ नहीं देय थे</span>. 1961 में, मिल्ग्राम में सामाजिक मनोविज्ञान <span style="color:red">में ए पीएचडी प्राप्त</span>. वह पर येल आस−पास वही समय एक सहायक प्रोफ़ेसर बन गया.<br>**CL**cov`ELemMS:` हार्वर्ड से मनोविज्ञान का मास्टर्स के लिए गया प्राथमिक अनुप्रयोगों <span style="color:red">के लिए इनकार दिया गया</span> आया, लेकिन अंत में उन्हें प्रवेश प्राप्त. प्रारंभिक बाधाओं का क्योंकि वह है मिल्ग्राम द्वारा क्वींस कॉलेज में मनोविज्ञान का कोई अवर <span style="color:green">पाठ्यक्रम लिया नहीं था</span>. 1961 में मिल्ग्राम उन्हें सामाजिक मनोविज्ञान में <span style="color:green">पीएचडी प्राप्त हो गया</span>. आम तौर पर जो उसी दौरान वह चिल्लाना विश्वविद्यालय में सहायक प्रोफ़ेसर हो गया. | CL`cov`ELemMS shows relatively more natural SOV Hindi word order, whereas L`cov`ELemMS appears scrambled and nonsensical. Note that both are semantically inadequate and unusable. |

*Continued on next page*

| Approach | Direction | Sample | Comment |
|---|---|---|---|
| CL$^{cov}$ELemMS | eng→ | **Input:** The members of Exodus company quickly dispersed. They were a part of the 2nd Shocktrooper Division, 2nd Army. They were originally two separate companies, Cohort and Dagger, before being consolidated. They were a cracked company, and Cohren felt mixed feelings about being promoted to 1st Lieutenant, mainly due to the reasons.<br>**Reference:** एज्जोडस कंपनी के सभी सैनिक तितर–बितर हो गए और सामान चेक करने लगे. यह कंपनी सेकंड आर्मी के सेकंड शॉकटूपर डिवीजन का हिस्सा थी. शुरुआत में यह कंपनी दो अलग–अलग कंपनियों में बंटी थी– कोहर्ट और डैगर. बाद में इसे मिलाकर एक कंपनी कर दिया गया था. इस कंपनी ने काफी कुछ झेला है. आज के हालात देखते हुए, कोहरेन भी फर्स्ट लेफ्टिनेंट बनाए जाने से बहुत ज्यादा खुश नहीं है.<br>L$^{cov}$ELemMS: पलायन कंपनी के सदस्यों जल्दी से तितर थे. वे 2 शॉकटूपर विभाजन का, 2 सेना का ६ भाग थे. वे मौलिक रूप से दो अलग कंपनियों, समूह और खंजर, पहले प्राणी समेकित थे. वे ए फटा हुआ कंपनी थे, और <span style="color:red">कोहरेन के बारे में प्राणी promoted को 1 लेफ्टिनेंट का मिश्रित भावनाएं अनुभव किया थे</span>, मुख्य रूप से देय को कारण ही.<br>CL$^{cov}$ELemMS: पलायन कंपनी का सदस्य जल्द ही बिखरा हुआ गया. वह दूसरा सेना का सेकंड शॉकटूपर विभाजन का भाग था. वह मौलिक रूप से दो स्वतंत्र कंपनियों थे, समूह और खंजर, एक साथ करने में पहुंचने से पहले. <span style="color:green">वह एक नामित (मसालेदार) कंपनी था</span>, और <span style="color:green">पहला एक लेफ्टिनेंट के रूप में पदोन्नति प्राप्त करने के लिए कोहरेन का मन में कम्पोजिट भावनाएं थे</span>, मुख्य रूप से वह प्रमोशन का कारणों से. | CL$^{cov}$ELemMS shows relatively more natural SOV Hindi word order, whereas L$^{cov}$ELemMS appears scrambled and nonsensical. Note that both are semantically inadequate and unusable. |
| CL$^{cov}$ELemMS | eng→ | **Input:** The number of complaints to authorities in Ukraine of a person speaking in Russian increased by 30% last year. Kiev actively encourages the population to report such cases in a bid to allegedly "protect" the country from Russian influence.<br>**Reference:** Le nombre de plaintes enregistrées auprès des autorités d'Ukraine pour signaler une personne parlant russe ont augmenté de 30 % l'année dernière. Kiev encourage activement la population à signaler de tels cas afin, selon elle, de "protéger" le pays contre l'influence russe.<br>L$^{cov}$ELemMS: Le nombre de plaintes à autorités dans Ukraine de <span style="color:red">un personne parlant dans Russe</span> augmenté par 30% <span style="color:red">dernier année</span>. Kiev activement encourage le population à rapport tel cas dans un offre à prétendument 'protéger' le pays depuis <span style="color:red">Russe influence</span>.<br>CL$^{cov}$ELemMS: Il nombre de plaintes en vue de autorités de Ukraine à propos <span style="color:green">un personne que parle russe</span> augmenter un 30% <span style="color:green">il année passé</span>. Kyiv encourager activement à le population à rapport ces cas dans un essayé de soi-disant "protéger" au pays de le <span style="color:green">influence russe</span>. | Several improvements in local phrase ordering, especially with noun-adjective ordering in French. |
| CL$^{cov}$ELemMS | eng→ | **Input:** He just started vacuuming again. I hope it's a good sign.<br>**Reference:** Il vient de recommencer à passer l'aspirateur. J'espère que c'est bon signe.<br>L$^{cov}$ELemMS: <span style="color:green">Il</span> juste commencé passer l'aspirateur encore. <span style="color:green">Je</span> espoir c'est un signe bien.<br>CL$^{cov}$ELemMS: <span style="color:red">Simplement</span> commencé à passer le aspirateur de nouveau. <span style="color:red">Attendez</span> que être un bien signe. | Spanish exhibits pro-drop, dropping the subject of the sentence. This is ungrammatical in French and results in a much worse translation than the original English-French translation. |

*Continued on next page*

| Approach | Direction | Sample | Comment |
|---|---|---|---|
| CL$^{cov}$ELemMS | eng→ | **Input:** High-street bakery chain Greggs has said that in the year ahead it does not plan to hike prices, and is aiming to open 160 branches as it targets locations outside city centres.<br>**Reference:** La chaîne de boulangerie au détail Greggs a déclaré que pour l'année à venir, elle ne prévoyait pas d'augmenter ses prix, mais envisageait d'ouvrir 160 filiales en périphérie des centres-villes.<br>**L$^{cov}$ELemMS:** Chaîne boulangerie rue principale Greggs a dit que dans le année devant il fait pas plan <span style="color:red">à randonnée les prix</span>, et est visée à ouvrir 160 succursales comme il cibles emplacements <span style="color:red">dehors ville centres</span>.<br>**CL$^{cov}$ELemMS:** Le chaîne de boulangeries Greggs ha adage que dans il suivant année non plans <span style="color:green">augmenter le prix</span> et a comme but ouvrir 160 branches, déjà que indiquer à emplacements dehors <span style="color:green">de le centres de le villes</span>. | The word "hike" in English here means to increase in the context of prices; while L$^{cov}$ELemMS get misled by a literal translation, using a cascade sidesteps this problem. This is because it is translated into the Spanish equivalent for "increase", which can safety be literally translated into French. This occurs several times since French is idiomatically closer to Spanish than English. |

# I Task details, prompts, and results

**Languages** For our task-based experiments, we study Czech, German, French, Hindi, Marathi, Spanish, Telugu, Vietnamese, and Turkish. `XNLI` contains 6 of our studied languages, including German, French, Hindi, Spanish, Turkish, and Vietnamese. `XStoryCloze` contains 3 languages, including Hindi, Spanish and Telugu, and `MMLU-ProX` contains 8 languages, including Czech, German, French, Hindi, Marathi, Spanish, Telugu, and Vietnamese.

**Prompts** See task prompt templates in Table 7. Materials are presented to the model in the manner as for MT, i.e. word meanings for each input word, and exemplars for the task, with a task-specific prompt.

**Results** See results for all tasks in Figure 8. Findings for these tasks are similar to those discussed in § 8, with LE generally performing the best of current approaches directly prompting the model to perform the task on the ciphered language input, and the cascaded system outperforming direct inference on the ciphered language.

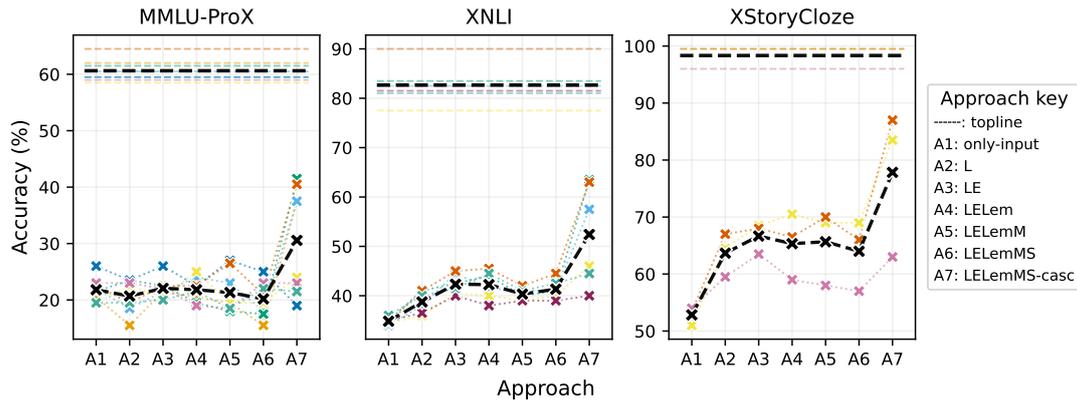

Figure 8: Results for all tasks with `GPT-5.1`.

---

**MMLU-Prox**

---

```
Answer the following multiple-choice question in Serra. Respond only with the numeric label of
the correct option.
Input:
{question,options}

...

Consider all the information provided above. Respond with *only* the output.
Input:
{question,options}
```

---

**XNLI**

---

```
You are given a premise and a hypothesis written in Serra. Determine whether the hypothesis is
entailed by, contradicted by, or neutral with respect to the premise. For entailment return `0`;
for neutral return `1`; for contradiction return `2`. Respond only with the numeric label of the
correct option.
Input:
{input}

...

Consider all the information provided above. Respond with *only* the output.
Input:
{input}
```

---

**XStoryCloze**

---

```
You are given a story, followed by two possible continuations, written in Serra. Choose the
continuation that best fits the prior context. Respond only with the numeric label of the
correct option.
Input:
{story,continuations}

...

Consider all the information provided above. Respond with *only* the output.
Input:
{story,continuations}
```

---

Table 7: Task prompts for Spanish. Materials for various approaches are included in the same manner as for MT, shown in Table 5.